# Quantum-Boosted High-Fidelity Deep Learning

Feng-ao Wang[1,3,†], Shaobo Chen[2,†], Yao Xuan[2,†], Junwei Liu[1,†], Qi Gao[2], Hongdong Zhu[2], Junjie Hou[2], Lixin Yuan[2], Jinyu Cheng[1,6], Chenxin Yi[1,5], Hai Wei[2], Yin Ma[2], Tao Xu[1,4*], Kai Wen[2,*], Yixue Li[1,3,4,7,8,9,*]

[1]Guangzhou National Laboratory, Guangzhou, 510005, China

[2]Beijing QBoson Quantum Technology Co., Ltd.

[3]Key Laboratory of Systems Health Science of Zhejiang Province, School of Life Science, Hangzhou Institute for Advanced Study, University of Chinese Academy of Sciences, Hangzhou, 310024, China

[4]GZMU-GIBH Joint School of Life Sciences, The Guangdong-Hong Kong-Macau Joint Laboratory for Cell Fate Regulation and Diseases, Guangzhou Medical University, Guangzhou, 511436, China.

[5]School of Intelligent Systems Engineering, Sun Yat-sen University, Shenzhen, 518107, China.

[6]The Chinese University of Hong Kong.

[7]School of Life Sciences and Biotechnology, Shanghai Jiao Tong University, Shanghai, 200240, China.

[8]Shanghai Institute of Nutrition and Health, Chinese Academy of Sciences, Shanghai, 200030, China.

[9]Bioland Laboratory, Guangzhou 510005, China.

† Feng-ao Wang, Shaobo Chen, Yao Xuan, Junwei Liu contributed equally to this work.

*Corresponding Author:

Tao Xu: xu_tao@gzlab.ac.cn

Kai Wen: wenk@boseq.com

Yixue Li: li_yixue@gzlab.ac.cn

**Abstract**

A fundamental limitation of probabilistic deep learning is its predominant reliance on Gaussian priors. This simplistic assumption prevents models from accurately capturing the complex, non-Gaussian landscapes of natural data, particularly in demanding domains like complex biological data, severely hindering the fidelity of the model for scientific discovery. The physically-grounded Boltzmann distribution offers a more expressive alternative, but it is computationally intractable on classical computers. To date, quantum approaches have been hampered by the insufficient qubit scale and operational stability required for the iterative demands of deep learning. Here, we bridge this gap by introducing the Quantum Boltzmann Machine-Variational Autoencoder (QBM-VAE), a large-scale and long-time stable hybrid quantum-classical architecture. Our framework leverages a quantum processor for efficient sampling from the Boltzmann distribution, enabling its use as a powerful prior within a deep generative model. Applied to million-scale single-cell datasets from multiple sources, the QBM-VAE generates a latent space that better preserves complex biological structures, consistently outperforming conventional Gaussian-based deep learning models like VAE and SCVI in essential tasks such as omics data integration, cell-type classification, and trajectory inference. It also provides a typical example of introducing a physics priori into deep learning to drive the model to acquire scientific discovery capabilities that breaks through data limitations. This work provides the demonstration of a practical quantum advantage in deep learning on a large-scale scientific problem and offers a transferable blueprint for developing hybrid quantum AI models.

**Introduction**

Deep learning frameworks achieve enhanced robustness to out-of-distribution inputs through the integration of probabilistic modeling, which enables systematic uncertainty quantification and reliable detection of unfamiliar data patterns[1,2]. When deep learning models explicitly incorporate probabilistic components, Gaussian priors are predominantly employed, particularly within variational inference frameworks[3] and Bayesian neural network architectures[4]. However, the stringent assumption of independence and identical distribution (IID) inherent to the Gaussian distribution constrains its capacity for data description[5]. In real-world scenarios such as biology, chemistry, and materials science, the probability distributions of samples exhibit far greater complexity and multifaceted characteristics compared to manually constructed datasets like MNIST and CIFAR-10[6-9]. The substantial discrepancy between the assumed prior distributions and the actual data distribution introduces significant biases into models trained on natural-world datasets, such as single-cell omics data[10] in different biological and disease background, fundamentally limiting their performance and increasing risks of invalidating theoretical guarantees due to biased or inconsistent parameter estimation-factors that collectively circumscribe the theoretical bounds of Gaussian distribution-based prior assumptions[11].

To enhance the fidelity of deep learning models in complex scenarios, refining the prior distribution assumptions of training data is imperative[12]. The Boltzmann distribution, a foundational construct in statistical physics, provides a principled framework for modeling probability distributions governed by energy landscapes[13,14]. This will have strong modeling capabilities and better adaptability for data generated in the natural world. However, practical implementation in large-scale systems is impeded by the computational intractability of the partition function $Z$[15], which necessitates summation over an exponentially large state space. Traditional sampling methods, such as Monte Carlo techniques[16], struggle to efficiently explore these high-dimensional landscapes, highlighting the critical need for novel computational methodologies to bridge theoretical rigor with practical feasibility[17].

As a compelling computational paradigm, quantum computing has been validated for its efficacy in addressing complex problems and enabling large-scale sampling[18,19]. Regrettably, constrained by the current quantum hardware in terms of computational qubit resources and its stringent operating conditions, large-scale real-world applications cannot yet be realized, making it difficult to meet the requirements of high-frequency and long-duration iterative computations in deep learning[20,21].

To address these challenges and validate the effect of optimized distributional priors, we propose a quantum-enhanced deep learning framework, the Quantum Boltzmann Machine-Variational Autoencoder (QBM-VAE). This model leverages our quantum hardware's inherent advantages in large-scale complex distribution sampling and long-time continuous operation to construct a hybrid quantum-classical computational architecture, enabling efficient training and inference. To demonstrate the superiority of Boltzmann-inspired latent variable representations via Gaussian based autoencoders, we applied QBM-VAE to multiple single-cell datasets in the order of millions, including PBMC, Immune Cell Atlas, Pancreas, Human Lung Cell Atlas, Human Fetal Lung Cell Atlas, and Human trabecular meshwork and ciliary body cell atlas collections. All of these datasets as the paradigmatic example of high-dimensional, non-Gaussian biological complexity. Comprehensive evaluations across multiple benchmarks, including single-cell integration, cell type classification, and lineage trajectory inference, consistently demonstrated that QBM-VAE's latent space preserves higher-order biological structures compared to conventional Gaussian-based variational autoencoders. These results underscore the potential of quantum-inspired statistical priors to resolve longstanding challenges in capturing the hierarchical organization and energy-driven dynamics inherent in biological systems. The proposed quantum-classical hybrid approach can be seamlessly transferred to the pre-training of large models to enhance the performance. This will mark a monumental breakthrough in the application of quantum computing and the evolution of artificial intelligence.

# Main

## The hybrid quantum-classical architecture of QBM-VAE

To enhance latent space modeling capability, we substitute the conventional Gaussian prior assumption in the variational autoencoder (VAE) model[22] with a Boltzmann distribution prior (Fig. 1a). This led to our proposal of QBM-VAE, an end-to-end self-supervised classical-quantum hybrid deep learning framework (Fig. 1b). Building upon the standard VAE model, we first enable discrete latent spaces modeling via reparameterization, then introduce a Boltzmann Machine (BM)[23], optimizing the Kullback-Leibler (KL) divergence[24] between the posterior distribution and a Boltzmann prior distribution derived from Boltzmann sampling. This imposes a rigorous Boltzmann constraint on the discretized latent variables. To enable model training, we established a classical-quantum hybrid computing architecture facilitating communication between classical and quantum processors (Fig. 1c). Within this framework, model parameters are transmitted to the quantum hardware for Boltzmann sampling. The resultant quantum samples are returned to the classical system, enabling end-to-end forward computation and gradient-based optimization via backpropagation[25].

This architecture achieves high-dimensional data mapping onto binarized latent features. Crucially, these features inherently conform to a Boltzmann prior distribution, overcoming limitations of conventional Gaussian priors and thus endowing the model with superior physical representational capacity for complex data manifolds. To benchmark the model's performance on real-world datasets, we applied our model to multiple single-cell omics datasets, aiming to better align with the underlying biological process within these single cells (Fig. 1d). Our result demonstrates enhanced feature extraction capabilities for these complex data distributions across diverse downstream analytical tasks.

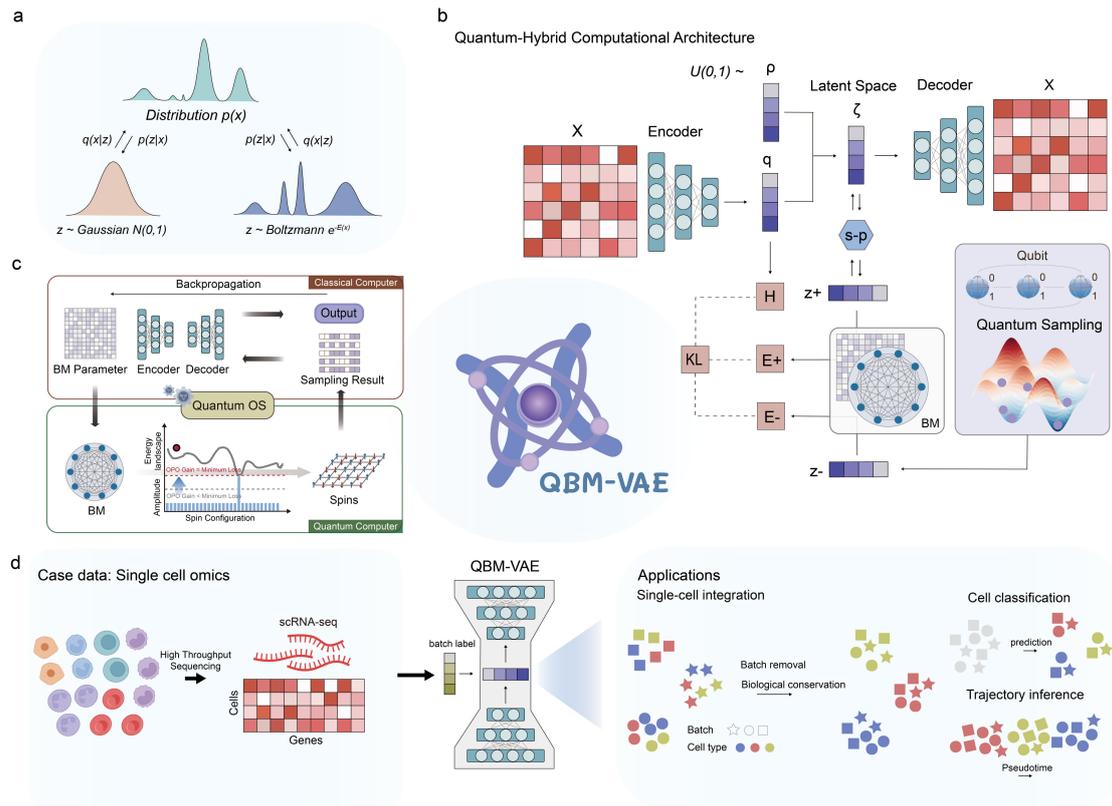

**Fig.1 The architecture of quantum-hybrid computation framework**.

**a.** Schematic of distribution approximation. A complex distribution is fitted using Gaussian and Boltzmann distributions.

**b.** Model architecture of QBM-VAE. Boltzmann distribution constraints are applied to the VAE latent space via the Boltzmann Machine (BM), with negative sampling implemented through quantum sampling.

**c.** Schematic of the classical-quantum hybrid computing system. Model parameters are mapped from the classical computer to the quantum processor via the quantum control interface. Photons evolve within the system, measuring quantum bit spin states as binary vectors returned to the classical computer.

**d.** Case application of QBM-VAE to single-cell omics. The QBM-VAE learns powerful latent representations from single-cell data that are applied to downstream analyses, including single-cell integration, cell type classification and trajectory inference.

**Large-scale stable quantum sampling enables QBM-VAE training**

Currently, the number of computable qubits available in different quantum hardware varies from several tens to several hundreds[26,27]. Such limited qubit counts and their non-fully connected property severely restrict the scope of their applications, making it challenging to meet the requirements for integration with real-world deep learning models[28,29]. In our quantum-hybrid computing framework, under the improvement of temperature control and vibration isolation system, our CIM hardware is capable of achieving continuous and stable solving for at least 12 hours with thousands of fully connected spins, and provides high-efficiency quantum sampling support for fast iterative training of our QBM-VAE model. Firstly, we construct a Boltzmann Machine (BM) based on the Ising model and implement Boltzmann sampling through a Coherent Ising Machine (CIM) and verify the ability of Boltzmann sampling in a 1000-node BM (Fig.2a).

Deep learning model training requires sustained high-precision floating-point operations and high-dimensional vector embedding. Our quantum hardware delivers persistent operational stability for extended sampling durations. Fig. 2c demonstrates continuous CIM operation solving a MAX-CUT problem of a 1000-node Möbius ladder graph (more details in Methods). The device consistently achieves the theoretical best solution (1498) with 99.8% accuracy over 12 hours, both its solving performance and duration of continuous stable solving far surpass those of current reported quantum hardware, validating robust stability. We further benchmarked quantum sampling against classical simulated annealing (Fig. 2d). Our quantum hardware significantly outperforms classical computation in solving minimum Hamiltonian problems across node scales (33-1025 nodes). This performance gap widens with increasing complexity, demonstrating the scalability advantage of our quantum sampling approach. Based on above results, our hybrid computing framework is capable of sustaining continuous, stable, long-duration operation and can implement Boltzmann sampling to facilitate the rapid training of QBM-VAE.

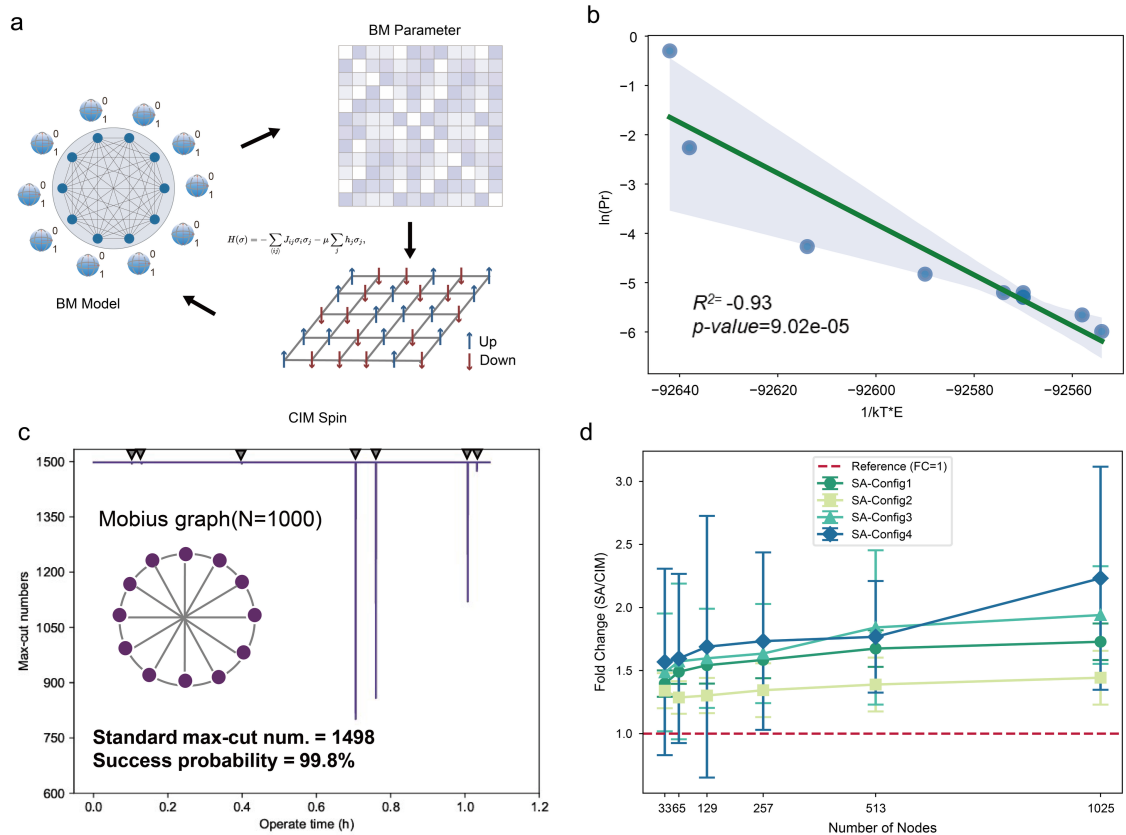

**Fig.2 Large-scale stable quantum sampling enables QBM-VAE training iterations.**

**a.** The left panel illustrates a schematic diagram of the natural mapping between the Boltzmann Machine (BM) and the Coherent Ising Machine (CIM). BM nodes represent spin qubits within the CIM.

**b.** Validation result of quantum Boltzmann sampling of BM parameters for a 1,024-spin system on the CIM. Each point represents one solution. The solution probability (log scale) shows a significant negative correlation (r² = -0.93, *p*-value = 9.02 × 10⁻⁵) with the scaled energy (1/kT × E).

**c.** Max-cut number of 1,000-node Möbius ladder graph (known optimum: 1498) solving by quantum-hybrid framework more than 1.2h.

**d.** Single-sample sampling time between the CIM and Simulated Annealing (SA) on a classical computer within 33, 65, 257, 513, and 1,025 spins. Diverse shapes correspond to distinct parameter configurations for simulated annealing.

## The benchmark of single cell omics integration

Single-cell omics, enabled by high-throughput sequencing, has transformed molecular cell biology by providing unprecedented resolution of cellular heterogeneity. To establish the superiority of QBM-VAE's Boltzmann-based prior, we benchmarked QBM-VAE on the critical task of single-cell omics data analysis. We rigorously compared its integration performance against state-of-the-art methods, including scVI, AUTOZI, LDVAE, and scPoli, as well as a standard VAE baseline that shared an identical architecture. Critically, all competing models rely on a conventional VAE framework with a standard Gaussian prior, allowing for a direct test of Boltzmann-based latent space.

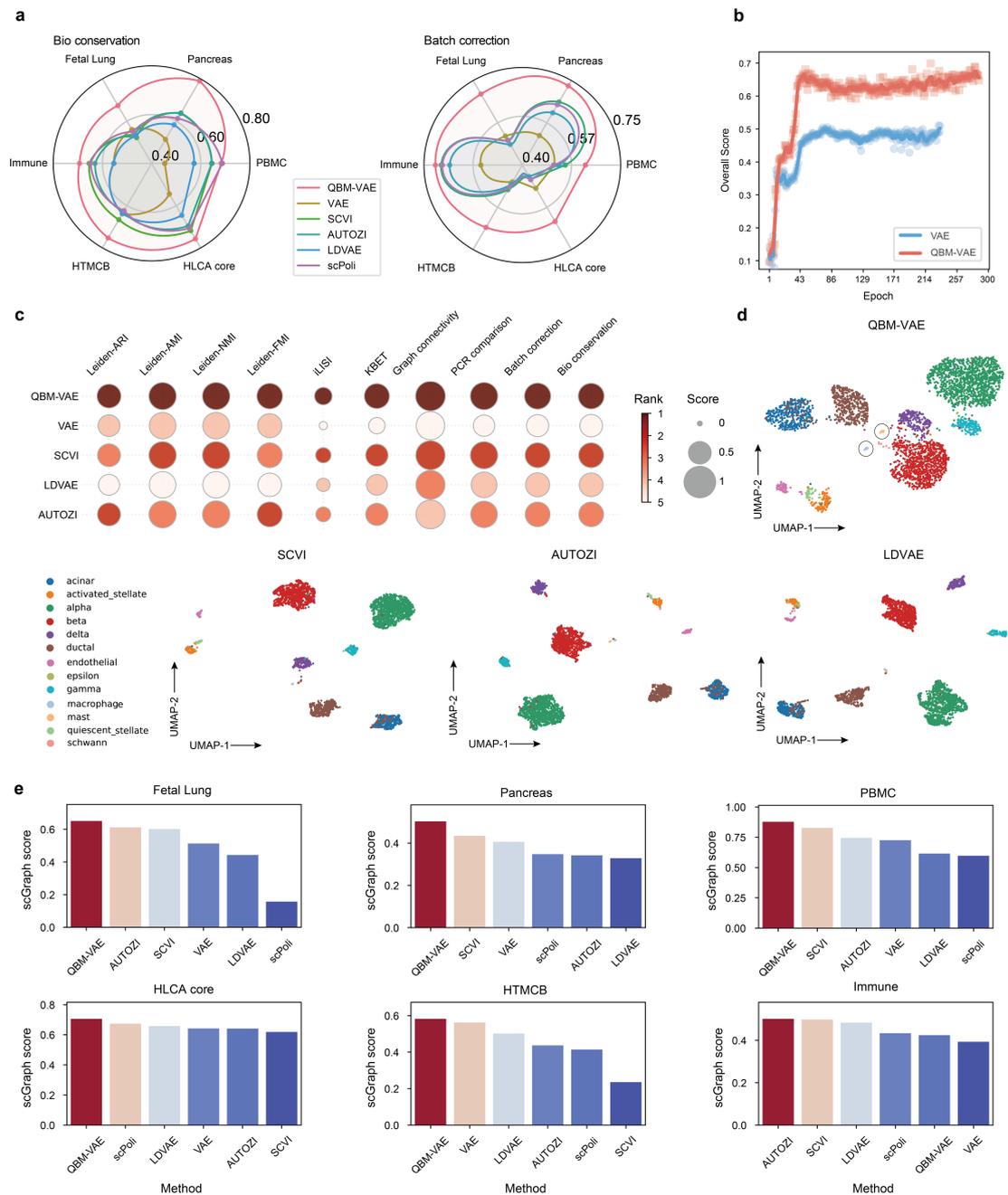

**Fig.3 The benchmark of single-cell omics integration.**

**a.** Radar plots summarizing the overall performance of QBM-VAE against competing methods across all six benchmark datasets. Performance is broken down into bio conservation (left) and batch correction (right) scores, calculated using the scIB framework. Higher scores (further from the center) indicate better performance.

**b.** Epoch-by-epoch comparison of the overall integration score between QBM-VAE and a standard, identically-parameterized VAE on the pancreas dataset.

**c.** Performance comparison on the pancreas dataset. The dot plot (left) shows normalized scores for individual metrics, grouped by biological conservation and batch correction. Dot size and color intensity correspond to the score.

**d.** The UMAP visualizations (right) compare the latent spaces generated by QBM-VAE and three competing methods. Cells are colored by their annotated cell type (legend provided at far left). QBM-VAE demonstrates superior separation of cell populations, notably resolving rare cell types like macrophages and mast cells into distinct clusters that are intermingled by other methods.

**e.** Bar plots showing the scGraph score for each method across the six individual datasets, demonstrating the consistent top-tier performance of QBM-VAE.

For this benchmark, we utilized six large-scale datasets comprising over one million cells, including the PBMC12k, Immune Cell Atlas, Pancreas, Human Lung Cell Atlas, Human Fetal Lung Cell Atlas, and Human trabecular meshwork and ciliary body cell atlas collections. Following the established scIB framework, we assessed the performance of each model against two primary objectives: the correction of batch effects and the preservation of biological variance. Biological variance was quantified using established cell clustering metrics (Adjusted Rand Index [ARI], Adjusted Mutual Information [AMI], Normalized Mutual Information [NMI], and Fowlkes-Mallows Index [FMI]) on Leiden-clustered latent embeddings. Batch effect correction was evaluated via the isolated-label LISI (iLISI) score, graph connectivity, and Principal Component Regression (PCR). An overall score for each objective was subsequently derived by integrating the respective metrics.

Our comprehensive benchmarking revealed that QBM-VAE consistently outperformed all baseline methods in both biological conservation and batch correction across all datasets (Fig. 3a), demonstrating its superior integration capabilities. To analyze the training dynamics, we tracked the integration performance of QBM-VAE against an identically parameterized VAE at each epoch on the pancreas dataset (Fig. 3a). The results reveal a stark difference in convergence. QBM-VAE demonstrated rapid performance gains, achieving a high integration score after only 50 epochs. At this early stage, it already substantially outperformed the standard VAE, which required much longer training to reach its subsequent, suboptimal peak. This advantage was particularly pronounced in the complex pancreas dataset (Fig. 3c). Visualization of the latent embeddings via UMAP further substantiated these findings. QBM-VAE uniquely resolved rare immune populations, such as macrophages and mast cells, into distinct clusters. In contrast, these cell types remained intermingled and poorly resolved within the latent spaces generated by scVI, LDVAE, and other competing methods (Fig. 3d).

**Boltzmann latent space preserves biologically meaningful topology**

Recent work has highlighted the limitations of evaluating latent space performance solely on cell clustering and batch removal metrics, as these can overlook the preservation of continuous, biologically meaningful structures. To address this, the scGraph metric was proposed to assess the conservation of topological structures within the latent space. We employed this metric to further evaluate QBM-VAE against other methods. The results show that QBM-VAE achieved a 20-30% improvement in the scGraph score across all tested datasets except the immune cell dataset (Fig. 3e). The performance of QBM-VAE across all test datasets demonstrated the robustness. This demonstrates that the latent space structured by the Boltzmann prior more faithfully recapitulates the intrinsic biological topology of the data compared to those based on a conventional Gaussian prior, thereby underscoring the superiority of our approach.

**Enhanced performance in downstream applications**

To further assess the utility of the learned representations, we applied the QBM-VAE embeddings to downstream analytical tasks, including cell-type classification and trajectory inference. For classification, we trained an XGBoost model using a 5-fold cross-validation scheme on the latent representations from the Human Lung Cell Atlas (HLCA). QBM-VAE achieved superior performance compared to all other methods(SCVI, AUTOZI and LDVAE) across metrics of accuracy (ACC), precision, recall, and F1 score (Fig. 4a).

Analysis of the confusion matrices for QBM-VAE and scVI was particularly revealing. QBM-VAE demonstrated a significantly better ability to distinguish between closely related subtypes (Fig. 4b), such as bronchial and alveolar fibroblasts. In contrast, scVI frequently misclassified bronchial fibroblasts as alveolar fibroblasts (Fig. 4c). Similarly, scVI failed to distinguish between classical and non-classical monocytes, whereas QBM-VAE resolved these populations effectively (Fig. 4c). These results highlight that the Boltzmann-informed latent space more accurately reflects the true cellular manifold.

Finally, we evaluated the preservation of developmental trajectories within the QBM-VAE latent space. Following the scIB framework, we analyzed the hematopoietic differentiation trajectory from hematopoietic stem and progenitor cells (HSPCs) to erythrocytes within the immune dataset. QBM-VAE achieved a trajectory conservation score of 0.995, surpassing scVI's score of 0.993 (Fig. 4d). This result demonstrates the meaningful biological architecture of the QBM-VAE latent space and its superior ability to preserve complex biological processes.

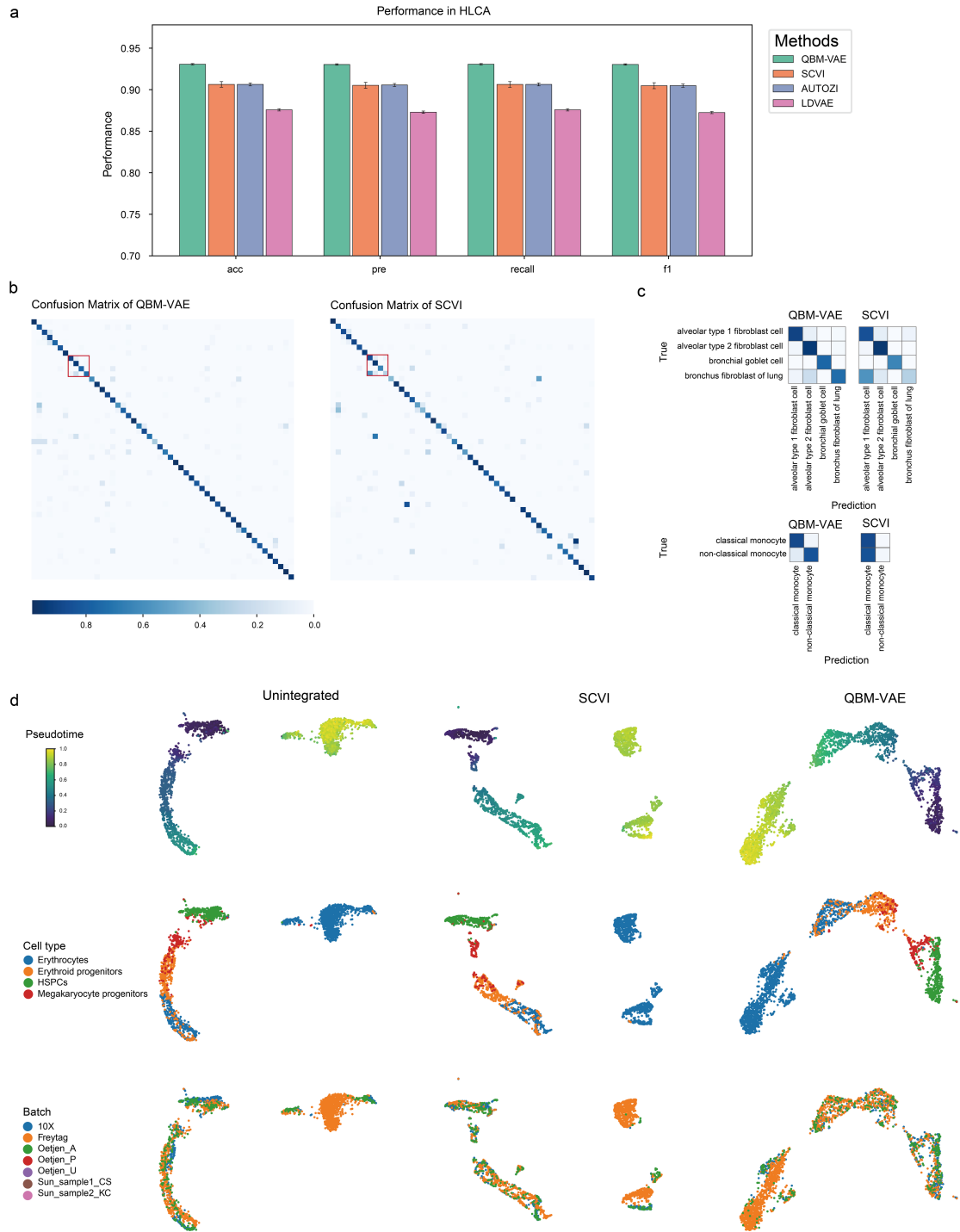

**Fig.4 QBM-VAE learns biologically superior representations for downstream applications.**

**a.** Comparison of cell-type classification performance on the Human Lung Cell Atlas (HLCA) dataset. Bar plots show the accuracy (ACC), precision (pre), recall, and F1 score achieved by an XGBoost classifier trained on the latent embeddings from QBM-VAE and competing methods. Error bars represent the standard deviation from a 5-

fold cross-validation.

**b.** Overall confusion matrices for cell-type classification using embeddings from QBM-VAE (left) and scVI (right). The stronger diagonal in the QBM-VAE matrix indicates higher accuracy. Red boxes highlight subtypes that are frequently confused by scVI but not QBM-VAE, which are detailed in (c).

**c.** Zoomed-in confusion matrices for closely related subtypes. QBM-VAE correctly distinguishes between fibroblast subtypes (top) and classical/non-classical monocyte subtypes (bottom), whereas scVI exhibits significant misclassification between these populations.

**d.** Visualization of the hematopoietic differentiation trajectory within the immune dataset. UMAP embeddings are shown for unintegrated data and for data integrated by scVI and QBM-VAE. Embeddings are colored by pseudotime score (top), cell type annotation (middle), and experimental batch (bottom). QBM-VAE preserves a more continuous developmental trajectory from hematopoietic stem and progenitor cells to erythrocytes while effectively correcting for batch effects.

**Discussion**

Deep generative models have achieved remarkable success, yet their capacity to model the natural world is fundamentally constrained by simplistic, mathematically convenient priors. For natural systems governed by statistical physics, we contend that this reliance on Gaussian assumptions introduces critical biases, limiting model fidelity. The Boltzmann distribution, which directly links probability to a physical energy landscape, offers a far more veridical foundation. However, its adoption has been historically thwarted by the classical intractability of sampling from it at scale. Here, we overcome this long-standing barrier by introducing a hybrid quantum-classical framework. By mapping the Ising model onto a Coherent Ising Machine (CIM), we perform large-scale Boltzmann sampling to train a deep generative model end-to-end, replacing the ubiquitous Gaussian prior with a physically meaningful, energy-based landscape.

To validate our approach, we selected the notoriously challenging domain of single-cell omics analysis. The high dimensionality, inherent noise, and complex structure of single-cell data have consistently challenged conventional deep learning models. By replacing the Gaussian prior in a variational autoencoder with our quantum-sampled Boltzmann distribution, we formulated the QBM-VAE. Our results provide unequivocal evidence of its superiority. Across multiple benchmarks, the QBM-VAE generated a more informative low-dimensional latent space that simultaneously enhanced denoising while preserving critical biological signals in single cell omics integration. In direct comparisons with established models such as standard autoencoders (VAE) and the domain leader, scVI and scPoli our method demonstrated markedly improved performance in downstream biological tasks. This leap in performance is directly attributable to the refined distributional prior, confirming that a physically-aware model yields more meaningful representations of complex biological systems.

The implications of this work extend far beyond single-cell biology and offer a new trajectory for the development of large-scale AI. The core challenge we addressed—efficiently sampling from complex, energy-based distributions—is a ubiquitous bottleneck in scientific AI. Our framework offers a tangible path to enrich the pre-training of next-generation foundation models. By enabling the direct integration of a sophisticated physical prior, our approach can imbue these models with a foundational physical inductive bias before task-specific fine-tuning. This promises to enhance model robustness and generalizability, potentially reducing the reliance on massive datasets. Furthermore, this work opens intriguing possibilities for a new class of generative models, including diffusion models. The reverse-denoising process, central to diffusion, could be guided by a physically grounded energy function sampled by our quantum processor, replacing or supplementing the standard Gaussian noise assumption. This could enable more efficient and structured generation of complex data with underlying physical constraints, such as molecules and materials.

While this study establishes a new paradigm, it also opens several avenues for future investigation. Our validation focused on single-cell omics; the immediate next step is to apply this framework to other domains where Boltzmann statistics are paramount, such as protein design and materials science, to demonstrate its generalizability. Moreover, while we demonstrate the empirical superiority of the Boltzmann-informed latent space, a deeper theoretical exploration of its properties is warranted. Future work should investigate the geometric and topological characteristics of this space to fully understand the mechanisms behind its enhanced performance and interpretability. With coherent Ising machines already exceeding 100,000 spins, these avenues represent not distant prospects but practical next steps to tackle problems of unprecedented scale and complexity.

In conclusion, we have demonstrated a powerful synergy between quantum computing and deep learning. By developing a practical method to replace the inadequate Gaussian assumption with a physically principled Boltzmann prior, we have fundamentally enhanced the ability of deep learning to model the natural world. This work establishes a robust blueprint for a new class of quantum-enhanced generative models. It marks a critical step towards a new paradigm of scientific discovery, where AI models, grounded in physical reality, become more insightful and reliable partners in exploring the complexities of our universe.

## Methods

In this study, a Variational Autoencoder (VAE) architecture with discrete spatial latent variables was employed. A reparameterization technique was utilized to achieve the transformation from continuous spatial variables $\zeta$ to a binary space. Subsequently, constraints were imposed on the transformed latent variables $z$. Specifically, $z$ was input into a Boltzmann Machine (BM), and the parameters of the BM were optimized to ensure that $z$ approximates the Boltzmann distribution as closely as possible. During the training process of the Boltzmann machine, a quantum annealer was used for model training.

## QBM-VAE architecture

### VAE module

For a generative model, its task is to learn a probability distribution $p_\theta(x)$, such that it approximates the true data distribution $p_{data}(x)$ as closely as possible. Here, $\theta$ denotes the parameters of the generative model, and the generative model can be expressed as:

$$p_\theta(x) = \int p_\theta(x|\zeta) p_\theta(\zeta) d\zeta$$

In a standard VAE model, where $x$ denotes the observed sample and $\zeta$ represents its latent-space features, a variational approximation $q_\varphi(\zeta|x)$ is introduced to approximate the true posterior $p_\theta(\zeta|x)$. The posterior distribution in VAE's variational inference can be expressed as:

$$\log p_\theta(x) = \log E_{q_\varphi(\zeta|x)} \left[ p_\theta(x|\zeta) p_\theta(\zeta) / q_\varphi(\zeta|x) \right]$$

We employ neural networks to parameterize $p_\theta(x|\zeta)$ and $q_\varphi(\zeta|x)$, corresponding to the decoder and encoder components, respectively. Two linear layers are used to project the input data into a 256-dimensional space, yielding a compressed representation $q$.

### Variational Inference in VAE

In standard VAEs, parameters $\theta$ and $\varphi$ are optimized to make the distributions $p_\theta$ and $q_\varphi$ approximate each other. According to Jensen's inequality, it holds that:

$$L(\theta, \varphi; x) = E_{q_\varphi(\zeta|x)} \left[ \log \left( p_\theta(x|\zeta) p_\theta(\zeta) / q_\varphi(\zeta|x) \right) \right] \leq \log p_\theta(x)$$

After transforming the formula, we obtain:

$$L(\theta, \varphi; x) = E_{q_\varphi(\zeta|x)}[\log p_\theta(x|\zeta)] - D_{KL}\left(q_\varphi(\zeta|x)|p_\theta(\zeta)\right)$$

The objective function $\mathcal{L}$ is the evidence lower bound (ELBO) of the VAE, and our goal is to maximize this ELBO. Since we introduce the binary variable $z$ through the transformation from $\zeta$ to $z$ via the function $r(\cdot)$, our optimization objective can be updated as:

$$L(\theta, \varphi, x) = E_{q_\varphi(\zeta|x)}[\log p_\theta(x|\zeta)] - D_{KL}\left(q_\varphi(z|x)|p_\theta(z)\right)$$

In classical VAEs, the prior is assumed to be a standard Gaussian distribution, which is a simple, symmetric, and fixed distribution. It assumes that the latent variables z are independent and identically distributed (i.i.d.). Typically, $p_\theta(\zeta) \sim N(0, I)$ is assumed, so the Optimization objective ELBO (Evidence Lower Bound) in standard VAEs can be written as:

$$\max_{\theta,\varphi} L(\theta, \varphi; x) = E_{q_\varphi(\zeta|x)}[\log p_\theta(x|\zeta)] - D_{KL}\left(q_\varphi(\zeta|x)|N(0, I)\right)$$

Therefore, in the standard VAE, the KL divergence can be specifically expressed as the following formula:

$$D_{KL}\left(q_\varphi(\zeta|x)|N(0, I)\right) = \frac{1}{2}\sum_{i=1}^{d}[\mu_i^2 + \sigma_i^2 - 1 - 2\log\sigma_i]$$

To enhance the capability of the latent space to model complex distributions, we introduce the Boltzmann distribution as the prior: Compared with the independence assumption of the standard normal distribution, the Boltzmann distribution can model richer latent space structures, such as multimodal distributions or nonlinear correlations, which are more consistent with real-world data distributions.

**Binary Reparameterization**

The CDF (Cumulative Distribution Function) represents an idealized approach to generating arbitrary samples from a uniform distribution, also known as inverse transform sampling. Therefore, we adopt the concept of inverse transform sampling to implement reparameterized sampling. We assume a CDF takes the following form:

$$F(\zeta) \equiv \int_0^\zeta q_\varphi(\zeta'|x)d\zeta'$$

The procedure for performing inverse transform sampling is as follows:

Sample a value $p$ from the uniform distribution $U(0,1)$.

Compute $\zeta = F^{-1}(p)$, where $F^{-1}$ denotes the inverse function of the CDF.

$\zeta$ is the sample drawn from the distribution $q_\phi(\zeta' \mid x)$.

By introducing a continuous relaxation variable $\zeta$ to approximate the discrete sampling process, gradient propagation is allowed, and a binary discrete variable $z$ is obtained. Specifically, a uniformly distributed variable $\rho$ is introduced, and $\zeta$ is jointly represented by the function $\zeta(\Phi, \rho)$. Further, $\zeta$ is transformed into $z$ through the function $r(\cdot)$, where the function $r(\cdot)$ employ spike-and-exponential transformation[30]:

$$r(\zeta_l \mid z_l = 0) = \delta(\zeta_l)$$

$$r(\zeta_l \mid z_l = 1) = \begin{cases} \beta \dfrac{e^{\beta \zeta_l}}{e^\beta - 1}, & if\ 0 < \zeta_l \le 1 \\ 0, & otherwise. \end{cases}$$

Here, we set $\beta = 0.5$.

Thus, we obtain the inverse transformation from $\zeta$ to $z$. We assume that each element in z follows a Bernoulli distribution, and the probability of $z$ can be output by the Encoder:

$$q_\phi(z_l = 1 \mid x) = q_l \quad \text{and} \quad q_\phi(z_l = 0 \mid x) = 1 - q_l$$

According to the inverse transformation of $r(\cdot)$, the cumulative distribution function (CDF) of $\zeta$ can be written in the following form at this point:

$$\rho_l = q_l \left( \frac{e^{\beta \zeta_l} - 1}{e^\beta - 1} \right) + (1 - q_l)$$

By performing inverse transform sampling on the above equation, the sampling form of $\zeta$ can be obtained.

$$\zeta_l(\rho_l, q_l) = \frac{1}{\beta} \log \left[ \frac{\max(\rho_l + q_l - 1, 0)}{q_l} (e^\beta - 1) + 1 \right]$$

**BM module**

Boltzmann machines are probabilistic models capable of representing complex multi-modal probability distributions. A Boltzmann Machine (BM) can be expressed as:

$$p_\theta(z) \equiv e^{-E_\theta(z)} / Z_\theta$$

$$Z_\theta \equiv \sum_z e^{-E_\theta(z)}$$

$$E_\theta(z) = \sum_l z_l h_l + \sum_{l<m} W_{lm} z_l z_m$$

For the binary latent variable $z$ we gained in the VAE module, we impose prior constraints based on the Boltzmann distribution using a BM. This is achieved by feeding $z$ into the BM and updating the BM parameters via KL calculation.

Building upon the above design, we have implemented an end-to-end QB-VAE framework that leverages BMs to enforce Boltzmann distribution constraints on the latent space of VAEs, thereby enhancing their representational capabilities.

In our model, the expectation term corresponds to the reconstruction loss. Notably, the KL divergence term can be expressed in the following form:

$$D_{KL}\left(q_{\varphi}(z|x)|p_{\theta}(z)\right) = E_{q_{\varphi}}[\log q_{\varphi}] - E_{q_{\varphi}}[\log p_{\theta}]$$

The entropy term can be written as:

$$\textbf{entropy term:}\ \ E_{q_{\varphi}}[\log q_{\varphi}] = -H(q_{\varphi})$$

$$H(q_{\varphi}) \equiv -E_{z \sim q_{\varphi}}[\log q_{\varphi}] = -E_{\rho \sim U}[\log q_{\varphi}(z(\rho, \varphi)|x)]$$

And the cross entropy term can be written as:

$$\textit{cross entropy term}: E_{q_{\varphi}}[\log p_{\theta}] = -H(q_{\varphi}, p_{\theta})$$

$$H(q_{\varphi}, p_{\theta}) = -E_{\rho \sim U}[\log p_{\theta}(z(\rho, \varphi))] = E_{\rho \sim U}[E_{\theta}(z(\rho, \varphi))] + \log Z_{\theta}$$

The calculation of the KL term is determined by the parameters of the BM, and the cross-entropy term is expressed by the energy of positive phase plus the energy of negative phase. The $\log Z_{\theta}$ term is implemented through CIM quantum sampling, Our goal is to minimize the reconstruction loss and the KL divergence term. During the optimization of the objective function, the transformed $z$ participates in the minimization of the KL divergence term, thereby imposing quantum-enhanced constraints on the pre-transformed $\zeta$.

**Quantum (CIM) sampling**

The sampling process during BM training can also be realized through quantum sampling. The process of constructing the Ising matrix from BM parameters is as follows:

---

**Algorithm 1 BM.__init__(n_visible, n_hidden)**

---

Input:

    n_visible – number of visible units

    n_hidden – number of hidden units

Output:

    An initialized BM model instance

1.    Compute n_unit ← n_visible + n_hidden
2.    // Initialize adjacency matrix
3.    Draw A_raw ∈ $\mathbb{R}^{n\_unit \times n\_unit}$ from $N(0,1)$
4.    Create mask M ← matrix of ones with zeros on diagonal
5.    A_masked ← A_raw ⊙ M
6.    A_sym ← (A_masked + A_masked$^T$) / 2
7.    Register A_sym as learnable parameter ad

8.    // Initialize bias vector
9.    Draw b_vec ∈ $\mathbb{R}^{n\_unit}$ from $N(0,1)$
10.   Register b_vec as learnable parameter b

11.   Precompute Ising matrix
12.   ising_matrix ← create_ising_matrix()

---

**Algorithm2 BM.create_ising_matrix()**

Input:

    Uses internal parameters ad ∈ ℝ^(n×n), b ∈ ℝ^n

Output:

    I ∈ ℝ^((n+1)×(n+1)), the Ising matrix

1. Let n ← dimension size of ad
2. Initialize I ∈ ℝ^((n+1)×(n+1)) as all zeros
3. Set I[0:n, 0:n] ← ad
4. Set I[0:n, n] ← b
5. Set I[n, 0:n] ← b
6. Zero out the diagonal of I
7. Return I as NumPy array

Subsequently, quantum annealing is employed to solve for the lowest Hamiltonian of the Ising matrix. Ising model can be written as:

$$H(\sigma) = -\sum_{i,j} J_{ij}\sigma_i\sigma_j - \mu \sum_i h_i \sigma_i$$

Where $\sigma$ is the discretized latent variable, the log partition function $LogZ$ can be approximately expressed as:

$$\log Z_\theta \approx \frac{1}{N} \sum_{n=1}^{N} E(h^T \sigma_n + \sigma_n^T W \sigma_n)$$

**Model Training**

The model is trained using a hybrid classical-quantum architecture with the following configurations:

**Optimization Strategy:** Both the Variational Autoencoder (VAE) and Boltzmann Machine (BM) modules are optimized using the Adam optimizer (Kingma & Ba, 2015). Fixed learning rates are set as $1 \times e^{-2}$ for the VAE module and $1 \times e^{-3}$ for the BM module. The reconstruction loss function is defined as the Mean Squared Error (MSE).

**Training Control:** An early stopping strategy is employed with a patience parameter of 10 epochs, monitoring the validation set ELBO.

**Computational Infrastructure**

**Classical Computing Unit:** Implemented using the PyTorch 2.0.0 framework (Paszke et al., 2019) with CUDA 11.8 acceleration on an NVIDIA RTX 3090 GPU.

**Quantum Computing Unit:** Quantum sampling processes are deployed on the QBoson Tiangong 550W photonic quantum computer.

**Hybrid Scheduling System:** Task scheduling and communication between classical and quantum components are managed by the Kaiwu SDK (v1.2)[31].

**Quantum sampling**

The sampling process during RBM training can also be realized through quantum sampling. We utilize quantum annealing for sampling on the TianGong Quantum Brain 550W and implement communication between classical and quantum computers according to the Kaiwu SDK. First, construct the Ising matrix based on the RBM parameters. The specific approach is as follows:

**Quantum-classical hybrid computing framework**

Here, we develop a large-scale quantum-classical hybrid computing framework, mainly composed of quantum hardware, coherent Ising machine, job scheduling system, and classical computing model. Gradient computation and parameter updating are performed on classical computation during the model iteration process. Quantum computer mainly generate samples from a Boltzmann distribution. In every training epoch, the Ising matrix $m$ is generated based on the BM parameters $\theta = \{w, h\}$, and then the energy distribution of the samples is generated by QC according to the Ising model.

**Setup of coherent Ising machine**

A continuous-wave (CW) laser light at a wavelength of 1550.1 nm is split into two lights using a coupler, and a portion of the coupler output is amplified by an erbium-doped fiber amplifier (EDFA1) and input to a balanced homodyne detector (BHD) as a local oscillator light. The other portion of the CW light is modulated into a pulse train whose pulse width and repetitive frequency are 30ps and 1 GHz, respectively, using an optical intensity modulator (IM1). The pulse train is then amplified by EDFA1 and separated into two paths with a coupler, where one of the outputs is used as injection pulses that convey the feedback signal from a field-programmable gate array (FPGA) system. The other portion is input into IM2, with which we control the pump amplitude for the signal and training pulses described in the main text. The pulses output from IM2 are again amplified, this time by EDFA2, and launched into a periodically poled lithium niobate (PPLN) waveguide module, where a 779.5-nm pulse train is generated via a second-harmonic generation (SHG). The SHG pulses are then input into another PPLN waveguide module that works as a phase-sensitive amplifier (PSA) in a fiber cavity.

**Long-time stability test of coherent Ising machine**

To confirm the computational stability of the entire quantum-classical hybrid computational framework, we performed the continuous solving max-cut of different Möbius graphs with different numbers of nodes 200, 400, 800. We performed 512 computations of the Max-Cut of different graph every ten seconds for 24 hours. The success probability refers to the ratio of analytical solutions acquired to total solving attempts.

**Single cell RNA-seq integration**

Integration performance was benchmarked following the scIB framework, which quantifies both biological conservation and batch-effect removal. Each dataset was integrated by our method and all baseline approaches, and the resulting low-dimensional embeddings were subjected to Leiden community detection to assign a cluster label to every cell. Bio-conservation scores and batch-removal scores were then computed by comparing these cluster labels with the reference cell-type annotations and known batch identifiers, respectively, according to the scIB metric suite.

**Batch effect removing**

Following previous studies on strategies for single-cell omics batch effect removal [], we applied one-hot embedding to represent the batch label. We then concatenated this batch embedding with the omics embedding for each cell before passing the combined representation to the decoder. In this way, the decoder's generative process is conditioned by the batch label, and the encoder's output embedding is trained to capture intrinsic biological representations rather than batch-specific factors.

**Cell type annotation**

To measure how well the integrated representations preserve biologically meaningful variation, we performed cell type annotation task using integrated embeddings. The whole dataset was randomly split into five-folds for cross-validation. In each fold, 80 % of the cells were used for training and hyper-parameter optimization of an XGBoost classifier, with the remaining 20 % held out for evaluation. The classifier was trained to predict the curated cell-type labels from the integrated embeddings. All performance metrics reported in the manuscript are averaged over the held-out test folds, ensuring an unbiased assessment of annotation accuracy.

**Trajectory inference**

Trajectory preservation was quantified with the scIB trajectory-inference benchmark. For each dataset, we first identified biologically coherent clusters (selected during preprocessing) and computed diffusion pseudotime (Scanpy, sc.tl.dpt) on the unintegrated counts within each batch; these within-batch trajectories were treated as the ground truth. After integration, diffusion pseudotime was re-computed on the joint neighborhood graph, but restricted to cells in its largest connected component. The root of each trajectory was set to the most extremal cell—in the space spanned by the first three diffusion components—belonging to the same cell-type cluster that contained the original root cells. Concordance between pre- and post-integration trajectories was measured by Spearman's rank correlation ($\rho$) of the pseudotime values. To aid interpretability, $\rho$ was linearly rescaled to the closed interval [0, 1], with 1 indicating perfect trajectory preservation.

**scRNA-seq datasets**

**Immune**. This dataset comprises 33,506 human immune cells (bone marrow and peripheral blood mononuclear cells; PBMCs) from 10 donors, representing 16 cell types. Data were aggregated from five studies employing multiple sequencing platforms.

**Pancreas**. This dataset contains 16,382 pancreatic cells from 9 batches, annotated into 14 distinct cell types.

**Human Lung Cell atlas (HLCA)**. The core HLCA dataset provides an integrated reference of the human respiratory system (lung parenchyma, respiratory airways, and nose), containing 584,944 lung cells from 166 samples across 107 individuals. It features detailed multi-level cell-type annotations and comprehensive metadata.

**Human fetal lung cell atlas**. Derived from a multiomic analysis of human fetal lungs (5-22 post-conception weeks), this atlas includes scRNA-seq profiles consisting of 71,752 cells from 29 batches (12 donors) across 8 developmental stages. It encompasses 14 broad cell types and 144 newly classified finer-grained cell types.

**Human trabecular meshwork and ciliary body cell atlas** This ocular dataset contains scRNA-seq profiles of 332,995 cells from the trabecular meshwork (TM) and ciliary body (CB) of 32 donors. It represents 15 broad cell types and spans a wide age range from newborn (0-28 days) to 85 years.

**scRNA-seq Data Preprocessing**

Single-cell RNA-seq data were processed in Scanpy. After standard quality control—removing cells with few detected genes or UMIs and discarding genes expressed in too few cells—the retained raw count matrix was used in two ways. For our model, we first normalised each cell to 10,000 total counts, applied a natural-log transformation [log (count + 1)], and then selected the 4,000 most highly variable genes (HVGs); these normalized, log-transformed HVGs constituted the input. For comparison methods implemented in scvi-tools (e.g. scVI, AUTOZI), we supplied the same set of 4,000 HVGs but kept them as raw counts, in line with the default requirements of those models.

**Baseline methods**

To evaluate the performance of our proposed method, we compared it against several established Variational Autoencoder (VAE) with Gaussian distribution as prior-based approaches for single-cell data integration. These included scVI, autoZI, and LDVAE (Linear Decoder VAE). Additionally, we implemented a custom VAE model with the same architecture analogous to that described for RBM-VAE for comparative purposes. For scVI, autoZI, and LDVAE, model architectures and training hyperparameters were based on the default configurations provided within the scvi-tools library.

**scVI** Single-cell Variational Inference (scVI) is a deep generative model that leverages VAEs and hierarchical Bayesian modeling to normalize and integrate single-cell RNA sequencing (scRNA-seq) data. It accounts for technical variability, including batch effects and library size differences, to learn a latent representation of cellular states.

**AUTOZI** An extension of scVI that incorporates an explicit model for zero-inflation in scRNA-seq data. It adaptively determines whether observed zeros correspond to biological absence of expression or technical dropouts, aiming for more accurate gene expression modeling.

**LDVAE** A VAE variant that employs a linear decoder. This simplification, compared to scVI's non-linear decoder, facilitates easier interpretation of the latent space dimensions and can be more robust for certain datasets or when interpretability is prioritized.

**Evaluation Metrics**

**Adjusted Rand Index (ARI)** The Rand index compares the overlap of two clusterings; it considers both correct clustering overlaps while also counting correct disagreements between two clusterings. We compared the cell-type labels with the NMI-optimized Leiden clustering computed on the integrated dataset. The adjustment of the Rand index corrects for randomly correct labels. An ARI of 0 or 1 corresponds to random labeling or a perfect match, respectively. We also used the scikit-learn implementation of the ARI.

**Adjusted Mutual Information (AMI)** Mutual information quantifies the amount of shared information between two partitions. AMI corrects this quantity for chance agreement, yielding values between 0 (no more agreement than expected by random label assignments) and 1 (perfect concordance). We computed AMI with the sklearn.metrics.

**Normalized Mutual Information (NMI)** The mutual information between two clustering is normalized by the geometric mean of their entropies, producing a score in the interval [0, 1]. A value of 1 indicates identical partitions, while 0 signifies independence. NMI was obtained with sklearn.metrics.

**Fowlkes-Mallows Index (FMI)** FMI is the geometric mean of precision and recall calculated on the pairwise contingency table of two clustering. It ranges from 0 (no pairwise agreement) to 1 (perfect pairwise agreement) and was calculated via sklearn.metrics.

**iLISI** iLISI measures batch mixing at the single-cell level. For each cell, the diversity of batch labels among its k nearest neighbours is quantified with the inverse Simpson index; the result is then inverted and averaged. Higher values indicate better batch mixing (i.e. stronger batch-effect removal). We used the implementation provided in the scIB metric suite.

**KNET** KNET evaluates biological conservation by computing the Shannon entropy of cell-type labels within each cell's k-nearest-neighbour set. Lower entropy (after appropriate scaling) implies better separation of distinct cell types. KNET scores were obtained with scIB.

**Graph connectivity** This metric assesses whether cells belonging to the same annotated cell type form a single connected component in the integrated neighbour graph. The score is the mean fraction of cells per cell type that lie in the largest component, ranging from 0 (fragmented) to 1 (fully connected). We followed the connectivity definition in scIB.

**PCR** comparison PCR quantifies the degree to which batch variables can be predicted from the integrated principal components. A lower coefficient of determination ($R^2$) indicates stronger batch removal. We computed PCR scores using linear regression on the first 50 PCs, as implemented in scIB.

**Accuracy (ACC)** The proportion of correctly predicted labels among all predictions: ACC = (TP + TN)/(TP + TN + FP + FN). Scores range from 0 to 1, with 1 denoting perfect classification.

**Precision** The fraction of true positives among all positive predictions: Precision = TP/(TP + FP). A value of 1 indicates no false-positive assignments. We used

sklearn.metrics.precision_score (macro-averaged across classes).

**Recall** The fraction of true positives retrieved among all actual positives: Recall = TP/(TP + FN). A value of 1 signifies that all relevant instances were recovered.

**F1** The harmonic mean of precision and recall: F1 = 2 × (Precision × Recall)/(Precision + Recall). It balances false positives and false negatives, attaining 1 for perfect predictions and 0 when either precision or recall is 0.

**Data availability**

All datasets analyzed in this study are publicly available: the immune-cell benchmark is hosted in the scIB reproducibility repository (https://github.com/theislab/scib-reproducibility); the pancreas dataset is available from the scPoli reproduce repository (https://github.com/theislab/scPoli_reproduce); bone-marrow mononuclear cells (BMMC) were obtained from the Gene Expression Omnibus under accession GSE194122; the Human Lung Cell Atlas core release can be downloaded from cellxgene (collection ID 6f6d381a-7701-4781-935c-db10d30de293); the human fetal lung atlas is available through the Wellcome Sanger Institute portal (https://fetal-lung.cellgeni.sanger.ac.uk/scRNA.html); and the human trabecular meshwork and ciliary body atlas is provided via cellxgene (collection ID 0cbf8ef8-87bd-4b51-8521-ec3f68183e11). No new primary data were generated for this study.